\documentclass[10pt,twocolumn]{article}

\usepackage[margin=0.75in]{geometry}

\usepackage[T1]{fontenc}
\usepackage[utf8]{inputenc}

\usepackage{amsmath,amssymb,amsfonts}

\usepackage{graphicx}
\usepackage{booktabs}
\usepackage{multirow}

\usepackage{algorithm}
\usepackage{algpseudocode}

\usepackage{xcolor}

\usepackage{cite}
\usepackage{url}
\usepackage[
    colorlinks=true,
    linkcolor=blue,
    citecolor=blue,
    urlcolor=blue
]{hyperref}

\usepackage{textcomp}

\title{HALLELUAI: A Hallucination-Aware AI System for Ultra-Realistic Image-to-Video Generation at Scale}

\author{
Aniket Sakpal\thanks{Expedia Group, Email: aniket17sakpal@gmail.com}
\and
Yang Jiang\thanks{Expedia Group, Email: jiang.yang.james@gmail.com}
\and
Rouzbeh Davoudi\thanks{Expedia Group, Email: rouzbeh.davoudi@gmail.com}
\and
Shayan Hassantabar\thanks{Expedia Group, Email: s\_hassantabar@yahoo.com}
\and
Mani Najmabadi\thanks{Expedia Group, Email: maninajmabadi@gmail.com}
}

\title{HALLELUAI: A Hallucination-Aware AI System for Ultra-Realistic Image-to-Video Generation at Scale}

\begin{document}

\maketitle

	\begin{abstract}
        AI-generated video is increasingly used across marketing, product storytelling, and creative workflows, yet automated; high-precision quality control remains a major constraint to scaling production. 
        We present HALLELUAI, an end-to-end system that moderates and regenerates image-to-video outputs to meet expert-level creative standards and deliver ultra-realistic videos with consistent end-user quality of experience (QoE) at scale. 
        The system integrates a video moderation module that evaluates frame-level aesthetics, temporal motion fidelity, and fine-grained hallucination risks relative to the source image, with an agentic regeneration module that iteratively fixes failures through prompt refinement, controlled camera adjustments, targeted model or image switching, and structured retry strategies. 
        The moderation logic is aligned with domain-specific creative guidelines and produces granular, machine-actionable feedback that directly drives regeneration. 
        In human-in-the-loop evaluations with creative experts, HALLELUAI shows strong alignment and reliably outputs ultra-realistic, production-grade videos suitable for product and marketing placements at scale. This framework advances trustworthy AI generated video content by enforcing visual realism, brand safety, and strict input-image fidelity while enabling image-to-video generation at scale. \\	 
	\end{abstract}  


    \maketitle

	\section{Introduction}
	\subsection{Motivation and Context}
    AI-generated video (AIGV) has rapidly become central to content production across visually intensive platforms such as digital advertising, social media, and product detail pages. Advances in diffusion-based image-to-video models now enable large-scale video synthesis at negligible marginal cost, supporting rapid experimentation, personalization, and localization. For domains such as travel, real estate, and e-commerce—where visual realism directly influences trust and conversion, AIGV offers a scalable alternative to traditional, resource-intensive production pipelines. 
    
    The economic incentives are substantial. Generative video systems compress production timelines from weeks to minutes and reduce per-asset costs by orders of magnitude relative to conventional workflows. These advantages position AIGV as a foundational technology for high-throughput visual storytelling. However, realizing this potential in production environments requires robust mechanisms to guarantee quality, realism, and strict fidelity to the source image. 

    \subsection{Problem Statement – Quality Challenges in Image-to-Video Generation }

    Despite rapid progress, image-to-video (I2V) models still produce artifacts that degrade quality of experience (QoE). Common failures include frame-level degradations (e.g., blur, noise, and brightness drift), temporal inconsistencies such as jitter and unnatural motion, and content hallucinations that introduce or alter scene elements relative to the source image \cite{cite13,cite12}.
    Hallucinations are particularly problematic because the input image serves as the ground truth. Deviations from it can misrepresent reality, reduce user trust, and create legal risks in domains such as travel and real estate \cite{cite12}.
    Current evaluation methods do not adequately address these challenges. Distributional metrics including Inception Score (IS) \cite{barratt2018note}, Fréchet Inception Distance (FID) \cite{heusel2017gans}, and Fréchet Video Distance (FVD) \cite{unterthiner2018towards} assess dataset-level realism but not fidelity to a specific input image. Likewise, existing video quality assessment methods often overlook AIGV-specific artifacts such as cross-frame inconsistencies and subtle hallucinations \cite{he2024cover, he2024videoscore, liu2024fvmd, wu2023exploring}.
    As a result, there is no reliable automated mechanism for detecting and mitigating these failures at scale. Recent experience with systems such as OpenAI's Sora \cite{cite15} further highlights the need for robust quality assessment and moderation frameworks for reliable I2V deployment \cite{cite9,cite10,cite14}.

\subsection{Gaps in Prior Work}
Despite recent advances in generative video evaluation, existing approaches remain insufficient for production-grade, image-conditioned AI-generated video.

\textbf{Per-asset, conditional evaluation:} Metrics such as IS \cite{barratt2018note}, FID \cite{heusel2017gans}, and FVD \cite{unterthiner2018towards} are designed for model-level comparison rather than asset-level acceptance. They do not condition on a specific input image and therefore fail to assess source-image fidelity, temporal stability, camera-motion quality, and hallucinations. FVD can also be unstable and may mis-rank models under limited sample sizes \cite{unterthiner2018towards}. Recent work further highlights the challenge of evaluating semantic consistency and multi-level structure in AI-generated video \cite{li2025multilevel}.

\textbf{Actionability:} Frameworks such as FVMD \cite{liu2024fvmd}, EvalCrafter \cite{liu2024evalcrafter}, VBench \cite{huang2023vbench}, and Video-Bench \cite{fu2023videobench} improve interpretability and alignment with human judgment, while VQAScore, GenAI-Bench, and VideoScore extend VQA-based evaluation to generative content \cite{wang2004ssim, lin2024evaluating, liu2024genaibench, he2024videoscore}. However, these methods remain primarily evaluative: they do not enforce source-image fidelity, gate individual assets, or provide diagnostics that directly support remediation.

\textbf{Hallucination detection:} Existing approaches focus on prompt consistency in text-to-video generation \cite{chu2024sora}, diffusion-model hallucinations \cite{aithal2024understanding}, or broader multimodal settings using uncertainty- and self-consistency-based techniques \cite{kadavath2022know, wang2022self}. Recent multimodal research likewise emphasizes the need for stronger grounding and semantic consistency \cite{zhang2025benchmarking, wang2025aigv}. However, these methods do not condition on a source image or detect fine-grained temporal and structural deviations.

\textbf{Closed-loop deployment:} Current benchmarks assess model capability but do not support per-asset gating, domain-specific compliance, or automated remediation. Consequently, the literature lacks end-to-end frameworks that combine expert-aligned evaluation, source-image fidelity, and iterative regeneration for deployment in high-trust domains such as travel and real estate \cite{zhang2025benchmarking, wang2025aigv}.

\subsection{Contributions}
This work introduces a production-oriented moderation and regeneration framework for image-to-video generation that addresses the above gaps.
\begin{itemize}
    \item \textbf{Per-asset, conditional video moderation:} a domain-aligned module evaluating each video relative to its input image across visual quality, temporal motion, and source-image fidelity.
    \item \textbf{Machine-actionable, expert-aligned diagnostics:} structured failure taxonomies, severity assessments, and pass/fail decisions mapping failure modes to corrective actions.
    \item \textbf{Fine-grained, temporally aware hallucination detection:} a layer identifying object-, structure-, and identity-level deviations, temporal inconsistencies, and artifacts relative to the source image.
    \item \textbf{Closed-loop, agentic regeneration:} a coupled module translating moderation feedback into targeted actions---prompt refinement, camera-control adjustment, model switching, base-image selection---iterating until quality criteria are met.
    \item \textbf{Human-in-the-loop validation:} an evaluation protocol demonstrating alignment with expert judgment and real-world deployability.
\end{itemize}

	\section{System Overview}
\begin{figure*}[t]\centering
    \includegraphics[width=\textwidth]{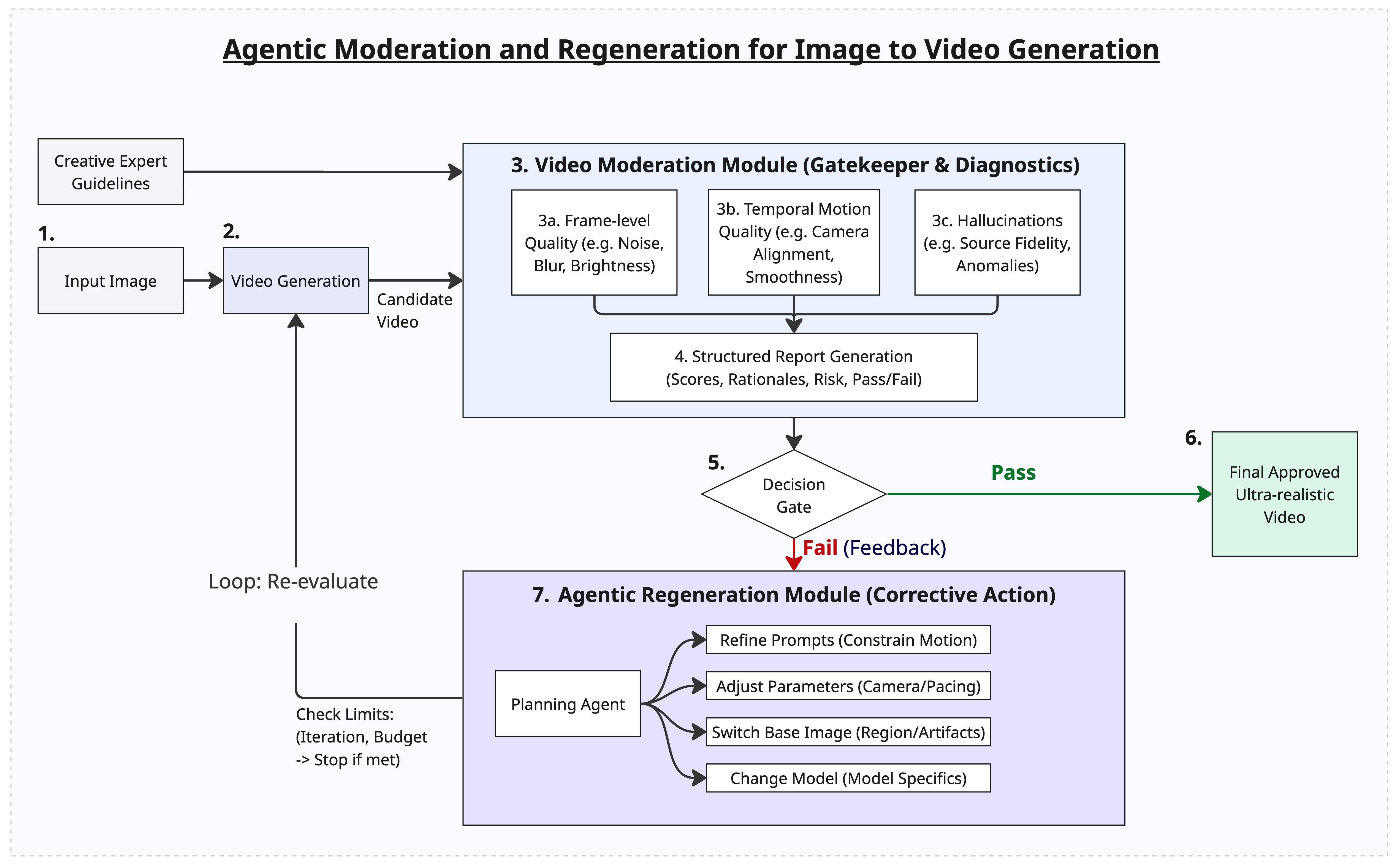}
    \caption{Architectural overview of the proposed closed-loop system for image-to-video generation. The framework integrates a diagnostic Video Moderation Module (3), calibrated by Creative Expert Guidelines, with an Agentic Regeneration Module (7). Candidate videos undergo iterative evaluation across frame-level, temporal, and fidelity dimensions (3a–c). Failures trigger a Planning Agent (7a) to execute targeted remediation strategies (7b–e) until the asset satisfies predefined quality thresholds or resource constraints are met.}\label{fig:system_design1}
    \end{figure*}

We propose an agentic closed-loop image-to-video generation system that unifies automated moderation with autonomous planning and targeted regeneration to enforce expert-defined creative quality and strict source-image fidelity at scale (Fig.~\ref{fig:system_design1}). Given an input image and creative guidelines (1), the system generates an initial video (2), which is evaluated by the Video Moderation Module (3) acting as both gatekeeper and diagnostics engine. The module assesses outputs across frame-level visual quality (3a), temporal motion quality (3b), and hallucination relative to the source image (3c), producing a structured moderation report (4) with dimension-wise scores, rationales, risk indicators, and a PASS/FAIL decision enforced by a decision gate (5).

For failed outputs, control is passed to the Agentic Regeneration Module (7), which converts feedback into autonomous actions. A planning agent (7a) maps failure categories and severities to targeted interventions, including prompt refinement for motion constraints (7b), camera direction or pacing adjustments (7c), base-image substitution for artifact-prone regions (7d), and alternative model selection for model-specific failures (7e). These actions drive targeted regeneration (8), after which outputs are re-evaluated by the moderation module.

The loop iterates until acceptance criteria are satisfied and final approval is reached (6), or until iteration, latency, or compute limits are exceeded (Fig.~\ref{fig:system_design1}). By coupling fine-grained diagnostics with autonomous planning and remediation, the system enables systematic quality control and scalable deployment of image-to-video generation.

    \section{Video Moderation Module}
The Video Moderation Module is the system's primary quality gate and diagnostic engine, assessing image-to-video outputs before approval or regeneration. For each candidate video, it evaluates domain-aligned creative and fidelity constraints across three dimensions: frame-level visual quality, temporal motion quality, and hallucination detection relative to the input image. The evaluation yields a structured, machine-actionable report with dimension-wise scores, calibrated risk levels, localized rationales, and a unified PASS/FAIL decision---designed to directly parameterize downstream regeneration for targeted correction rather than unguided retries. The module is grounded in a hierarchical problem taxonomy (Fig.~\ref{fig:system_design1}) categorizing common failure modes in ultra-realistic image-to-video generation, providing a shared abstraction layer that maps detected failures to corrective strategies such as prompt constraints, parameter adjustments, base-image substitution, or model switching.

  \subsection{Frame-Level Quality Signals: Blur, Contrast, Brightness, and Noise}
    We compute four complementary frame-level signals over a generated video $V=\{f_1,\dots,f_T\}$ and a reference input image $I$, with all frames and $I$ converted to grayscale for statistics. Each signal is summarized via normalized ratios and compared against calibrated thresholds to produce a unified PASS/FAIL decision with a concise reason.
    
    \paragraph{Blur (Sharpness Degradation).}
    Blur is quantified via the variance of the Laplacian: for each frame $f_t$, we compute a sharpness score $\mathcal{L}_t=\mathrm{Var}(\nabla^2 f_t)$, then measure the relative sharpness drop $\Delta_{\mathrm{blur}}$ by comparing the anchor $\mathcal{L}_1$ to the minimum sharpness across the video.
    
    \paragraph{Contrast (Low/High Contrast).}
    Contrast is quantified via the standard deviation of grayscale intensities. With reference $C_{\mathrm{ref}}=\sigma(I)$ and frame contrast $C_t=\sigma(f_t)$, we compute normalized decrease $\Delta_{\downarrow}$ and increase $\Delta_{\uparrow}$ relative to reference, optionally tracking within-video spread to capture excessive variability.
    
    \paragraph{Brightness/Exposure (Low/High Exposure).}
    Brightness is quantified via mean grayscale intensity. With $B_{\mathrm{ref}}=\mu(I)$ and $B_t=\mu(f_t)$, we compute normalized exposure decrease $\Delta^{B}_{\downarrow}$ and increase $\Delta^{B}_{\uparrow}$ from the minimum and maximum brightness relative to the reference.
    
    \paragraph{Noise (Stochastic Artifacts).}
    Noise is estimated via a per-frame Laplacian-variance score $N_t$; frames are marked noisy when $N_t$ exceeds a threshold $\tau_{\mathrm{noise}}$, and the video is flagged if the percentage of noisy frames exceeds a small tolerance (e.g., $5\%$), capturing persistent stochastic artifacts.
    
    \begin{algorithm}[H]
    \caption{Unified Frame-Level Quality Detection (Blur, Contrast, Brightness, Noise)}
    \label{alg:frame_quality_unified}
    \begin{algorithmic}
    \Require Reference image $I$, video $V=\{f_1,\dots,f_T\}$
    \Require Thresholds $\tau_{\mathrm{blur}}, \tau_{\downarrow}, \tau_{\uparrow}, \tau^{B}_{\downarrow}, \tau^{B}_{\uparrow}, \tau_{\mathrm{noise}}, \tau_{p}$
    \Ensure Decision $d \in \{\mathrm{PASS},\mathrm{FAIL}\}$, reason $r$
    
    \State $d\leftarrow\mathrm{PASS}$,\quad $r\leftarrow$ Acceptable Quality
    \State Convert $I$ to grayscale once and compute $C_{\mathrm{ref}}=\sigma(I)$,\; $B_{\mathrm{ref}}=\mu(I)$
    
    \Statex \hrulefill
    \State \textbf{(1) Blur / Sharpness}
    \State Compute $\mathcal{L}_t=\mathrm{Var}(\nabla^2 f_t)$ for all $t$
    \State $\Delta_{\mathrm{blur}} \leftarrow (\mathcal{L}_1-\min_t \mathcal{L}_t)/\mathcal{L}_1$
    \If{$\Delta_{\mathrm{blur}}>\tau_{\mathrm{blur}}$}
        \State $d\leftarrow\mathrm{FAIL}$;\; $r\leftarrow$ Sharpness Degradation
    \EndIf
    
    \Statex \hrulefill
    \State \textbf{(2) Contrast}
    \State Compute $C_t=\sigma(f_t)$ for all $t$
    \State $\Delta_{\downarrow} \leftarrow (C_{\mathrm{ref}}-\min_t C_t)/C_{\mathrm{ref}}$
    \State $\Delta_{\uparrow} \leftarrow (\max_t C_t-C_{\mathrm{ref}})/C_{\mathrm{ref}}$
    \If{$d=\mathrm{PASS}$ \textbf{and} $\Delta_{\downarrow}>\tau_{\downarrow}$}
        \State $d\leftarrow\mathrm{FAIL}$;\; $r\leftarrow$ Low Contrast
    \ElsIf{$d=\mathrm{PASS}$ \textbf{and} $\Delta_{\uparrow}>\tau_{\uparrow}$}
        \State $d\leftarrow\mathrm{FAIL}$;\; $r\leftarrow$ High Contrast
    \EndIf
    
    \Statex \hrulefill
    \State \textbf{(3) Brightness / Exposure}
    \State Compute $B_t=\mu(f_t)$ for all $t$
    \State $\Delta^{B}_{\downarrow} \leftarrow (B_{\mathrm{ref}}-\min_t B_t)/B_{\mathrm{ref}}$
    \State $\Delta^{B}_{\uparrow} \leftarrow (\max_t B_t-B_{\mathrm{ref}})/B_{\mathrm{ref}}$
    \If{$d=\mathrm{PASS}$ \textbf{and} $\Delta^{B}_{\downarrow}>\tau^{B}_{\downarrow}$}
        \State $d\leftarrow\mathrm{FAIL}$;\; $r\leftarrow$ Low Exposure
    \ElsIf{$d=\mathrm{PASS}$ \textbf{and} $\Delta^{B}_{\uparrow}>\tau^{B}_{\uparrow}$}
        \State $d\leftarrow\mathrm{FAIL}$;\; $r\leftarrow$ High Exposure
    \EndIf
    
    \Statex \hrulefill
    \State \textbf{(4) Noise}
    \State Compute per-frame noise score $N_t$ (Laplacian variance) and mark noisy frames where $N_t>\tau_{\mathrm{noise}}$
    \State $p_{\mathrm{noise}} \leftarrow \frac{1}{T}\sum_{t=1}^{T}\mathbb{I}[N_t>\tau_{\mathrm{noise}}]$
    \If{$d=\mathrm{PASS}$ \textbf{and} $p_{\mathrm{noise}}>\tau_{p}$}
        \State $d\leftarrow\mathrm{FAIL}$;\; $r\leftarrow$ Noise Artifacts
    \EndIf
    
    \end{algorithmic}
    \end{algorithm}

    \subsection{Temporal Motion Quality}
This component ensures camera motion is cinematic, intentional, and domain-appropriate. It evaluates (i) \textit{prompt alignment}—whether the dominant motion direction/style matches the prompt intent—and (ii) \textit{motion intensity and smoothness}—whether the pace is neither stagnant nor aggressive and free of jitter. We compute dense optical flow using RAFT to obtain a per-pixel displacement field between consecutive frames, deriving robust frame-level motion statistics that classify the video as \textit{Low Motion}, \textit{High Motion}, or \textit{Acceptable}, with additional jitter flags for high-frequency instability.

\subsubsection{Unpleasant Camera Motion (Intensity \& Smoothness) Module}
We estimate motion intensity from the RAFT flow magnitude. At each step $t$, RAFT produces a dense displacement field $\mathbf{u}_t(x)\in\mathbb{R}^2$ mapping pixels from $f_t$ to $f_{t+1}$. We compute the per-pixel magnitude $m_t(x)=\lVert \mathbf{u}_t(x)\rVert_2$ and summarize motion via a robust statistic: the mean magnitude of the top-$q$ fraction of pixels (e.g., $q=15\%$), $M_t=\mathrm{Mean}(\mathrm{Top}_q(\{m_t(x)\}))$, focusing on salient moving regions and reducing background-noise sensitivity. Video-level intensity is $\mu_M=\mathrm{Mean}_t(M_t)$ and stability $\sigma_M=\mathrm{Std}_t(M_t)$. The module flags \textit{Low Motion} when $\mu_M$ falls below a threshold (stagnant/slow pans), \textit{High Motion} when $\mu_M$ exceeds a threshold (whip-pan/aggressive moves), and \textit{Jitter} when $\sigma_M$ or the high-frequency variation of $M_t$ exceeds a threshold (abrupt accelerations). These outcomes map directly to actionable remediation (reduce motion strength, tighten camera range, or stabilize).

\subsubsection{Prompt Alignment Module}
This module verifies that observed camera motion matches the prompt-specified intent (e.g., pan left-to-right, tilt up, dolly-in/zoom-in). We estimate camera motion using TAPIR point tracking: salient points initialized on the first frame are tracked through the video to obtain 2D trajectories. To reduce object-motion confounds, we preferentially sample edge points (high-gradient regions) and aggregate trajectories into group-level displacement statistics. Letting $p_i^t\in\mathbb{R}^2$ denote tracked point $i$ at time $t$, we compute net displacement $\Delta p_i = p_i^T - p_i^1$ and summarize motion by the dominant direction $\hat{\mathbf{v}}=\mathrm{Normalize}(\mathrm{Median}_i(\Delta p_i))$. We further use structured point groups (e.g., left-edge set $\mathcal{P}_L$, top-edge set $\mathcal{P}_U$) to detect zoom/dolly: if $\mathcal{P}_L$ moves left while $\mathcal{P}_U$ moves up (divergence from the image center), the motion is consistent with zoom-in/dolly-in, whereas convergence indicates zoom-out/dolly-out. The observed label $\hat{y}$ is compared against the prompt intent $y_{\mathrm{prompt}}$ to yield an alignment score $s_{\mathrm{align}}$, triggering a FAIL when below threshold.

    \begin{algorithm}[H]
    \caption{Temporal Motion Quality: Intensity \& Smoothness + Prompt Intent Alignment (TAPIR)}
    \label{alg:temporal_motion_quality_tapir}
    \begin{algorithmic}
    \Require Video $V=\{f_1,\dots,f_T\}$, prompt motion intent $y_{\mathrm{prompt}}$
    \Require Top-pixel fraction $q$ (e.g., $0.15$)
    \Require Thresholds $\tau_{\mathrm{low}},\tau_{\mathrm{high}},\tau_{\mathrm{jit}}$, alignment threshold $\tau_{\mathrm{align}}$
    \Ensure Decision $d_{\mathrm{motion}}\in\{\mathrm{PASS},\mathrm{FAIL}\}$ and reason $r_{\mathrm{motion}}$
    
    \State $d_{\mathrm{motion}}\leftarrow\mathrm{PASS}$,\quad $r_{\mathrm{motion}}\leftarrow$ Acceptable Motion
    
    \Statex \hrulefill
    \State \textbf{(A) Unpleasant Camera Motion: Intensity \& Smoothness (RAFT)}
    \For{$t=1$ to $T-1$}
        \State $\mathbf{u}_t \leftarrow \mathrm{RAFT}(f_t,f_{t+1})$
        \State $m_t(x)\leftarrow \lVert \mathbf{u}_t(x)\rVert_2$ for all pixels $x$
        \State $M_t \leftarrow \mathrm{Mean}\big(\mathrm{Top}_q(\{m_t(x)\})\big)$
    \EndFor
    \State $\mu_M \leftarrow \mathrm{Mean}_t(M_t)$,\quad $\sigma_M \leftarrow \mathrm{Std}_t(M_t)$
    \If{$\mu_M < \tau_{\mathrm{low}}$}
        \State $d_{\mathrm{motion}}\leftarrow\mathrm{FAIL}$;\; $r_{\mathrm{motion}}\leftarrow$ Low Motion (Stagnant)
    \ElsIf{$\mu_M > \tau_{\mathrm{high}}$}
        \State $d_{\mathrm{motion}}\leftarrow\mathrm{FAIL}$;\; $r_{\mathrm{motion}}\leftarrow$ High Motion (Aggressive)
    \ElsIf{$\sigma_M > \tau_{\mathrm{jit}}$}
        \State $d_{\mathrm{motion}}\leftarrow\mathrm{FAIL}$;\; $r_{\mathrm{motion}}\leftarrow$ Jitter / Instability
    \EndIf
    
    \Statex \hrulefill
    \State \textbf{(B) Prompt Intent Alignment (TAPIR)}
    
    \State \textbf{(B1) Point Initialization \& Tracking:}
    \State Initialize an edge-biased point set $\mathcal{P}$ on the first frame $f_1$
    \State Track points $\{p_i^t\}_{t=1}^T$ using TAPIR and compute net displacements
    \Statex \quad $\Delta p_i \leftarrow p_i^T - p_i^1$
    
    \State \textbf{(B2) Dominant Motion Estimation:}
    \State Estimate global camera translation
    \Statex \quad $\hat{\mathbf{v}} \leftarrow \mathrm{Normalize}\!\left(\mathrm{Median}_{i\in\mathcal{P}}(\Delta p_i)\right)$
    
    \State \textbf{(B3) Zoom / Dolly Detection:}
    \State Partition points into spatial edge sets (e.g., left $\mathcal{P}_L$, top $\mathcal{P}_U$)
    \Statex \quad $\delta_L \leftarrow \mathrm{Median}_{i\in\mathcal{P}_L}(\Delta p_i)$,\;
    $\delta_U \leftarrow \mathrm{Median}_{i\in\mathcal{P}_U}(\Delta p_i)$
    
    \State \textbf{(B4) Motion Intent Classification:}
    \State Infer observed motion label
    \Statex \quad $\hat{y} \leftarrow \mathrm{ClassifyMotion}(\hat{\mathbf{v}}, \delta_L, \delta_U)$
    
    \State \textbf{(B5) Prompt Alignment Check:}
    \State Compute alignment score $s_{\mathrm{align}} \leftarrow \mathbb{I}[\hat{y}=y_{\mathrm{prompt}}]$
    \If{$d_{\mathrm{motion}}=\mathrm{PASS}$ \textbf{and} $s_{\mathrm{align}} < \tau_{\mathrm{align}}$}
        \State $d_{\mathrm{motion}}\leftarrow\mathrm{FAIL}$;\; $r_{\mathrm{motion}}\leftarrow$ Prompt Misalignment
    \EndIf
    
    \end{algorithmic}
    \end{algorithm}

    \subsection{Hallucinations}
We evaluate two types of hallucination in image-to-video generation: \emph{object hallucination} and \emph{new-structure hallucination}. Object hallucination refers to failures in preserving temporal consistency, physically plausible motion, and coherent object interactions across frames relative to the input image, leading to object drift, deformation, and other artifacts. \emph{New-structure hallucination} is the synthesis of objects or scene elements absent from the source image—particularly important in high-trust domains (e.g., travel, real estate), where introducing nonexistent structures may misrepresent reality and increase legal and reputational risk. The following section details the annotation categories for each type.

\begin{figure}[H]
    \centering
    \includegraphics[width=\columnwidth]{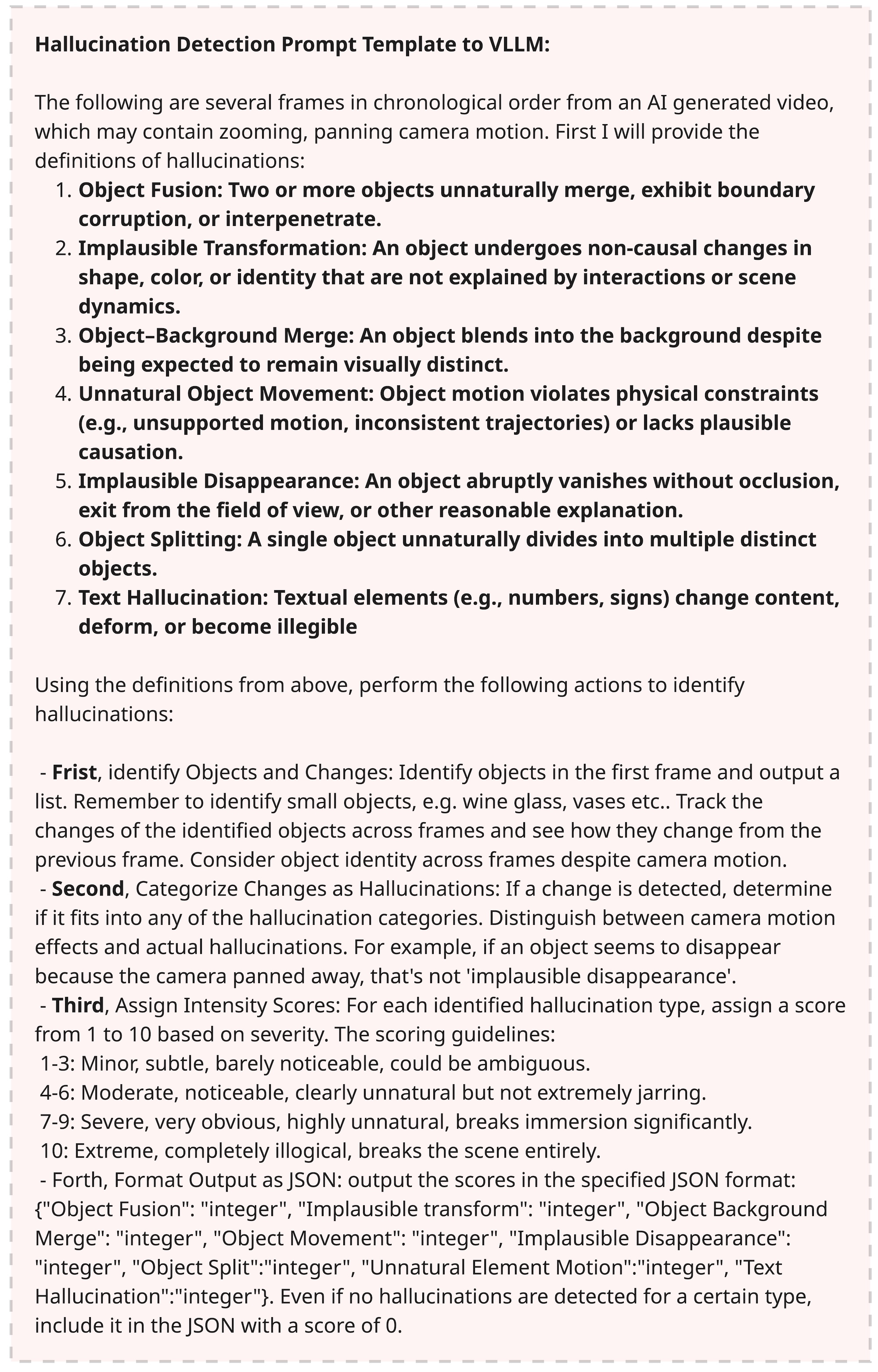}
    \caption{Prompt template provided to the VLLM for hallucination detection. It defines seven hallucination categories and instructs the model to identify and track objects across frames, categorize changes as hallucinations (distinguishing them from camera motion), assign a 1–10 severity score per category, and output results as JSON (0 if none detected).}
    \label{fig:my_image}
\end{figure}

\subsubsection{Hallucination Categories}
\begin{enumerate}
\item \textbf{Object Hallucination}
\begin{itemize}
\item \textbf{Object Fusion}: Two or more objects unnaturally merge, exhibit boundary corruption, or interpenetrate.
\item \textbf{Implausible Transformation}: An object undergoes non-causal changes in shape, color, or identity unexplained by interactions or scene dynamics.
\item \textbf{Object--Background Merge}: An object blends into the background despite being expected to remain visually distinct.
\item \textbf{Unnatural Object Movement}: Object motion violates physical constraints (e.g., unsupported motion, inconsistent trajectories) or lacks plausible causation.
\item \textbf{Implausible Disappearance}: An object abruptly vanishes without occlusion, exit from view, or other reasonable explanation.
\item \textbf{Object Splitting}: A single object unnaturally divides into multiple distinct objects.
\item \textbf{Text Hallucination}: Textual elements (e.g., numbers, signs) change content, deform, or become illegible.
\end{itemize}
\item \textbf{New Structure Hallucination:}
\begin{itemize}
\item \textbf{Natural entry}: Objects absent from the source image appear by entering the field of view with temporally continuous, physically plausible motion.
\item \textbf{Camera-induced entry}: Previously unseen objects become visible due to global camera motion (e.g., panning, tilting, zooming) rather than object motion.
\item \textbf{Abrupt emergence}: Objects not in the source image appear instantaneously without reasonable physical explanation.
\end{itemize}
\end{enumerate}

\subsubsection{Hallucination Detection Module}
\begin{figure*}[t]\centering
    \includegraphics[width=\textwidth]{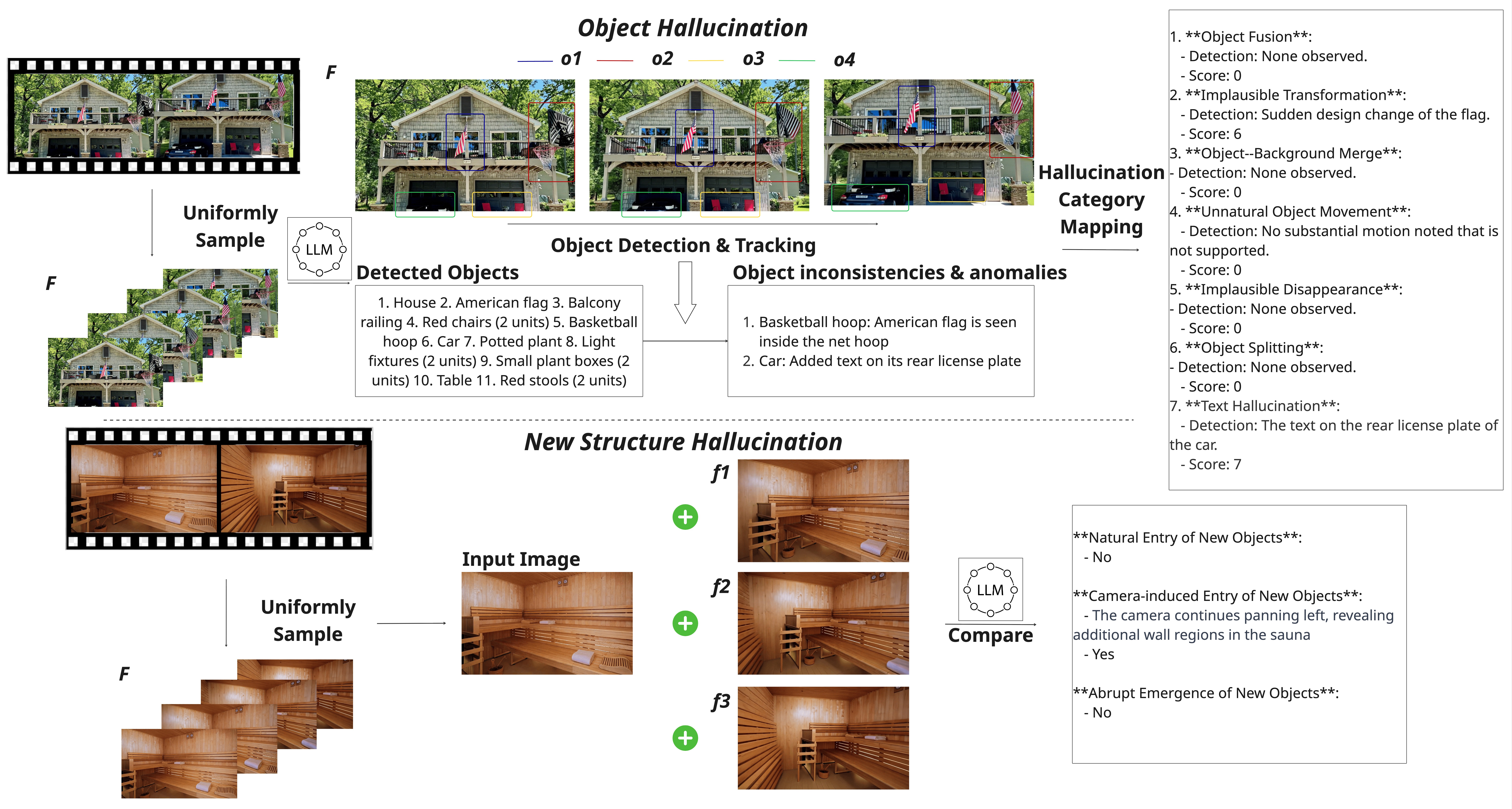}
    \caption{Hallucination detection module for Image2Video generation. Uniformly sampled frames are analyzed with an MLLM to (top) detect and track input-image objects, map temporal inconsistencies to predefined object-hallucination categories, and aggregate category scores for a final decision; and (bottom) compare each sampled frame with the input image to identify new-structure hallucinations (natural entry, camera-induced entry, or abrupt emergence)}\label{fig:system_design}
    \end{figure*}

As illustrated in Fig.~\ref{fig:system_design}, we propose a hallucination-detection module for videos generated by an Image2Video model. Given a generated video, we uniformly sample frames $F=\{f_1,\dots,f_n\}$. To detect object hallucinations, we first extract objects $O=\{o_1,\dots,o_m\}$ from the input image using a multimodal large language model (MLLM), then provide $O$, the input image, and $F$ to an MLLM (e.g., GPT-4o) to identify object-level inconsistencies across time, mapped to the predefined categories $C=\{c_1,\dots,c_m\}$. For each category $c_i$, the MLLM outputs a severity score $S^{i}_o \in [0,10]$ and a brief rationale when evidence is present, following the prompt template shown in Fig.~\ref{fig:my_image}. The object-hallucination decision $d_{\text{object}}$ aggregates the category-wise scores, thresholding each against $\tau_{\text{object}}^{i}$. To detect new-structure hallucinations, we compare each sampled frame $f_i$ with the input image to determine whether novel objects appear under three cases: (1) natural entry, (2) camera-induced entry, and (3) abrupt emergence. An MLLM produces a per-frame binary decision $d^{i}_{\text{new}}$, and the video-level decision is $d_{\text{new}} = \bigcup_{i=1}^{n} d^{i}_{\text{new}}$.
    
\subsection{Scoring and Risk Assessment}
\label{subsec:risk_assessment}
The module aggregates frame-level, temporal-consistency, and hallucination scores via domain-calibrated weights into a unified verdict, paired with a narrative summary contextualizing each issue against creative standards and Quality of Experience (QoE). A \texttt{PASS}/\texttt{FAIL} decision follows from thresholded expert-derived criteria, with a standardized risk level (\textit{No}, \textit{Low}, \textit{Medium}, \textit{High}) tied to specific failures. Outputs form a tuple of failure codes, severities, and remediation levers (e.g., reduce motion, adjust exposure, switch models), serving both as a gating mechanism and as input to the agentic regeneration module for targeted, iterative corrections.

\section{Agentic Regeneration Module}
The Agentic Regeneration Module converts moderation findings into targeted corrective actions via a closed-loop planner--executor process. A planning agent interprets frame-level, temporal, and hallucination signals relative to the source image, selects a minimal yet effective intervention set, and issues an updated generation request; the regenerated asset is re-scored, looping until quality thresholds or operational limits are met.

\paragraph{Inputs and Outputs.}
Inputs comprise the source image and metadata (e.g., campaign, channel), prior generation configuration (prompt, model, seed, motion parameters), and the structured moderation report (failure codes, severities, rationales, PASS/FAIL). Outputs comprise a next-generation plan (revised prompt, actions, parameters, chosen model or base image) and execution artifacts (regenerated video with a full action log for reproducibility).

\paragraph{Action Policy.}
The agent draws on a constrained intervention library: (i) prompt refinements enforcing motion clarity, object preservation, source-image fidelity, and aesthetic stability; (ii) camera/motion adjustments to reduce jitter, overshoot, or blur; (iii) base-image selection when artifacts cluster around occlusions or ambiguous regions; and (iv) model switching or tuning for model-specific failures. A cost- and efficacy-aware policy prioritizes low-cost, minimally invasive actions, escalates for high-severity or repeated failures, and bundles actions when interactions are likely.

\paragraph{Iteration and Governance.}
The loop (Generate $\rightarrow$ Moderate $\rightarrow$ Regenerate) terminates on PASS, budget exhaustion, or escalation to human review for persistent issues, logging all decisions for auditability. Coupling structured diagnostics with a cost-aware strategy, the module converts failing candidates into compliant assets while preserving input-image fidelity and minimizing manual oversight.
    
    \section{Tuning and Evaluation}
    
    Threshold calibration and parameter tuning were performed using 58 AI-generated video clips across the three moderation components. For hallucination detection, we used GPT-4o as the underlying LLM. Prompts were iteratively refined to elicit accurate reasoning for each hallucination category. Recall, defined as the proportion of hallucination-containing videos correctly detected, was used as the primary metric for selecting the final prompts. Category-specific hallucination thresholds were then optimized to maximize sensitivity for detecting high-risk videos, as defined in Section~\ref{subsec:risk_assessment}.  We then evaluate the proposed tuned system using a human-in-the-loop protocol designed to measure alignment with expert creative judgment and readiness for production deployment. The protocol consists of three stages: \textbf{Shadow Mode} (calibration), \textbf{Pseudo Production} (gating performance), and \textbf{Production QA} (longitudinal monitoring).
    
    \subsection{Stage 1: Shadow Mode}
    In Shadow Mode, unseen 146 AI-generated video clips were independently evaluated by creative expert reviewers, the HALLELUAI moderation system, and industry benchmark methods, without cross-visibility of decisions. Each clip was standardized through trimming and normalizing the aspect-ratio before evaluation.

    \subsubsection{Creative Expert Alignment}
    
    First, we evaluated the alignment of moderation systems by comparing their output to reviews from human creative experts on the video dataset. The HALLELUAI moderation system achieves 88\% precision and an agreement rate of 87\%. The results are summarized in Table~\ref{tab:ai_metrics}
    
    
    
    
    \paragraph{Agreement Analysis.}
    
    
    

    \begin{table}[t]
    \centering
    \small
    \caption{HALLELUAI moderation Performance Metrics Compared to Expert Decisions}
    \label{tab:ai_metrics}
    \begin{tabular}{lc}
    \toprule
    Metric & Value \\
    \midrule
    Overall PASS/FAIL Agreement & 86.9\% \\
    Precision on AI-approved assets & 88\% \\
    Recall on AI-approved assets & 74\% \\
    \bottomrule
    \end{tabular}
    \end{table}


    Precision is the key metric, as it directly reflects the quality and reliability of videos approved by the system.
    
    The system exhibits conservative gating behavior, filtering a portion of expert-approved clips while limiting false approvals.
    
    \paragraph{Stage-Level Diagnostic Breakdown}
    
    To analyze performance across moderation components, we decompose precision by category, as shown in Table~\ref{tab:stage_diagnostics}.
    

    \begin{table*}[!htbp]
    \centering
    \small
    \caption{Stage-level contribution to Precision and False Positives (FP). Percentages reflect the proportion of total observed errors attributable to each signal.}
    \label{tab:stage_diagnostics}
    \begin{tabular}{llc}
    \toprule
    \textbf{Module} & \textbf{Sub-Category} & \textbf{Precision} \\
    \midrule
    \multirow{4}{*}{Frame-Level Quality} 
     & Blur & 100\% \\
     & Contrast &  97\% \\
     & Brightness & 100\% \\
     & Noise & 100\% \\
    \midrule
    \multirow{2}{*}{Temporal Motion Quality} 
     & Prompt Alignment & 100\% \\
     & Motion Intensity &  97\% \\
    \midrule
    \multirow{3}{*}{Hallucination Detection} 
     & Object Hallucination &  97\% \\
     & New Structure Hallucination & 97\% \\
     & Text Hallucination & 100\% \\
    \bottomrule
    \end{tabular}
    \end{table*}

    The majority of precision-impacting errors arise from fine-grained object-level hallucinations. Motion intensity thresholds contribute primarily to false negatives, indicating conservative rejection behavior. Frame-level degradations contribute minimally to disagreement.
    \subsubsection{Benchmark Comparison}
Since no open-source systems provide comprehensive evaluation for image-to-video generation, we compared our system against UGC video quality assessment (VQA) models such as DOVER \cite{wu2023exploring} and COVER \cite{he2024cover}, as well as multimodal large language models (MLLMs) via zero-shot prompting (See Fig.~\ref{fig:Nova_prompt} for prompt details). The MLLM prompt was designed to evaluate the three components our system addresses---frame-level quality, temporal motion quality, and hallucination---with frames uniformly sampled at 2 fps.

As shown in Table~\ref{tab:performance_comparison}, UGC VQA models like DOVER and COVER could not differentiate PASS from FAIL videos, as they were not designed for image-to-video generation and lack hallucination-detection metrics. The MLLM-based method showed low precision, unable to detect artifacts in AI-generated videos---demonstrating that without detailed, comprehensive prompting, MLLMs cannot reliably identify such artifacts.

\begin{figure}[!htbp]
\centering
\includegraphics[width=\columnwidth]{Evaluation_prompt.jpg}
\caption{Evaluation Prompt Designed for MLLMs to Evaluate Frame-level Quality,Temporal Motion Quality and Hallucination in Image2Video Generation}
\label{fig:Nova_prompt}
\end{figure}

\begin{table}[!htbp]
\caption{Performance Comparison of Different Evaluation Methods}
\label{tab:performance_comparison}
\centering
\begin{tabular}{lcccc}
\toprule
Method & Prec. & Rec. & F1 & Acc. \\
\midrule
DOVER & 0.20 & 0.21 & 0.20 & 0.56 \\
COVER & 0.27 & 0.33 & 0.30 & 0.58 \\
Qwen3-VL-8B & 0.28 & 1.00 & 0.43 & 0.30 \\
NOVA 2 Lite & 0.25 & 0.74 & 0.37 & 0.34 \\
\textbf{HALLELUAI} & \textbf{0.88} & \textbf{0.74} & \textbf{0.81} & \textbf{0.87} \\
\bottomrule
\end{tabular}
\end{table}

\subsection{Stage 2: Pseudo Production (Gating Performance)}
    
    In Pseudo Production mode, we evaluate the combined moderation + regeneration system end-to-end by sending only system-approved clips for human creative review. Of 1,158 system-approved clips, 1,126 were also accepted by experts, yielding ~97\% precision and indicating that the vast majority of surfaced clips are production-ready. This precision is not directly comparable to Shadow Mode, since outputs here have undergone iterative regeneration and refinement before evaluation—the system is not merely filtering errors but actively improving clips, driving higher agreement with expert judgment.
        
    \subsection{Stage 3: Production QA (Longitudinal Monitoring)}
    
    Following calibration, ongoing monitoring employs random sampling (1–5\%) of system-approved assets for expert review. Precision levels of 97\% remain consistent with Pseudo Production Mode results, and regression testing ensures that threshold or regeneration-policy updates do not degrade previously validated performance. Deployed at scale, the system generated approximately 70,000+ videos, further validating its reliability.
    
    \subsection{Evaluation Summary}
    
    These results demonstrate that structured moderation combined with agentic regeneration can function as a reliable production-grade quality control layer for ultra-realistic image-conditioned video generation.
         
    \section{Conclusion}
HALLELUAI is an expert-aligned, production-oriented system that makes image-to-video quality assurance a controllable, closed-loop process, coupling per-asset moderation---aesthetics, motion quality, and source-image--conditioned hallucination detection---with an agentic regeneration policy that converts failures into targeted fixes rather than blind retries. Human-in-the-loop validation shows strong agreement with creative experts, supporting deployment as a high-precision gate for scalable AIGV. By enforcing input-image fidelity, emitting machine-actionable diagnostics, and maintaining auditable decision trails, it bridges benchmark-style evaluation and real-world creative governance---to our knowledge, the first integrated moderation-and-regeneration framework purpose-built for ultra-realistic, image-conditioned video generation at scale.

\bibliographystyle{plain}
\bibliography{references}
    


    \end{document}